\documentclass[conference]{IEEEtran}

\renewcommand\IEEEkeywordsname{Keywords}
\IEEEoverridecommandlockouts
\usepackage{cite}
\usepackage{amsmath,amssymb,amsfonts}
\usepackage{algorithmic}
\usepackage{graphicx}
\usepackage{textcomp}
\usepackage[svgnames]{xcolor}
\usepackage{float} 
\usepackage{lipsum}
\usepackage{tikz}
\usepackage{amsmath}
\usepackage{lipsum}
\usepackage{multirow}
\usepackage{url}

\newcommand\MyHead[2]{%
  \multicolumn{1}{|c|}{\parbox{#1}{\centering #2}}
}

\usetikzlibrary{shadows}
\usetikzlibrary{decorations.pathreplacing}

\def\BibTeX{{\rm B\kern-.05em{\sc i\kern-.025em b}\kern-.08em
    T\kern-.1667em\lower.7ex\hbox{E}\kern-.125emX}}

\begin{document}

\title{Unique Faces Recognition in Videos \\
}

\author{\IEEEauthorblockN{Jiahao Huo}
\IEEEauthorblockA{\textit{School of Computer Science and Applied Mathematics} \\
\textit{University of the Witwatersrand}\\
}
\and
\IEEEauthorblockN{ Terence L van Zyl}
\IEEEauthorblockA{\textit{WITS Institute of Data Science} \\
\textit{School of Computer Science and Applied Mathematics} \\
\textit{University of the Witwatersrand}\\
}
}

\maketitle

\begin{abstract}
This paper tackles face recognition in videos employing metric learning methods and similarity ranking models. The paper compares the use of the Siamese network with contrastive loss and Triplet Network with triplet loss implementing the following architectures: Google/Inception architecture, 3D Convolutional Network (C3D), and a 2-D Long short-term memory (LSTM) Recurrent Neural Network. We make use of still images and sequences from videos for training the networks and compare the performances implementing the above architectures. The dataset used was the YouTube Face Database designed for investigating the problem of face recognition in videos.
The contribution of this paper is two-fold: to begin, the experiments have established 3-D Convolutional networks and 2-D LSTMs with the contrastive loss on image sequences do not outperform Google/Inception architecture with contrastive loss in top $n$ rank face retrievals with still images. However, the 3-D Convolution networks and 2-D LSTM with triplet Loss outperform the Google/Inception with triplet loss in top $n$ rank face retrievals on the dataset; second, a Support Vector Machine (SVM) was used in conjunction with the CNNs' learned feature representations for facial identification. The results show that feature representation learned with triplet loss is significantly better for n-shot facial identification compared to contrastive loss. The most useful feature representations for facial identification are from the 2-D LSTM with triplet loss. The experiments show that learning spatio-temporal features from video sequences is beneficial for facial recognition in videos.
\end{abstract}


\section{Introduction}
There is an increase in demand for face recognition systems\cite{parkhi2015deep, sun2014deepjoint, sun2014deep}. A facial recognition model is a system trained with machine learning that can be used to identify or verify a person from images and videos. Identification is made by extracting and comparing the facial features of each unique individual \cite{krizhevsky2012imagenet,varior2016gated}. To the best of our knowledge, there has been limited research using similarity learning for facial recognition in videos with 3-D Convolutional Neural networks (CNNs) and Long Short Term Memory (LSTM) recurrent neural networks. In this research, we implement similarity learning functions for both still images and sequences of images (videos). We use the similarity learning to learn feature representations by predicting relative distances between the still images/sequences based on their similarity. Furthermore, we compare the feature representations employing a simple linear classifier to identify individuals. We used triplet loss and contrastive loss in metric learning techniques to train similarity learning models\cite{wang2014face,schroff2015facenet,yue2015beyond,tran2015learning}. We compared the face recognition performances between learning from still images of videos on a well-known Google/Inception architecture to short sequences of images learning with 3-D Convolutional networks and 2-D LSTM.
\\ \\ 
Video data capture both spatial and temporal information that can be interpreted by machines to complete a task like classification or identification. Previous research related to video-based facial recognition extracted frames (still images) from the videos and used these as inputs to CNNs~\cite{ding2017trunk,parkhi2015deep}. However, individual frames only contain spatial information in the 2-D plane and do not encapsulate temporal information across frames~\cite{rao2017attention,yang2017neural}. The additional temporal information may improve the accuracy of facial recognition. Further sequential frames allow us to capture multiple views of the faces, facial movements, and behaviors which are all features not exploited in a standard CNN trained on still images. We propose the use of data fusion and machine learning to capture both spatial and temporal information from videos when performing facial recognition. Data fusion is a way of fusing multiple data types (spatial and temporal in our case) before performing machine learning analysis and can either be done as early or late fusion. Early fusion consists of fusing raw or pre-processed data obtained from sensors before using it as input to a model \cite{khaleghi2013multisensor}. Late fusion uses separate models for separate data sources and only fuses at the prediction or decision-making level \cite{wu2016deep}. In our research, we make use of early fusion. Our choice is informed by the fact that videos are composed of still frames related to one another in time and that we can easily fuse the data by stacking sequential frames of $n$ size to gain access to spatio-temporal information for analysis with 3-D Convolutions and 2-D LSTMs.
\\ \\ 
In addressing the above motivation for data fusion across time, we proposed a version of a 3D Convolutional Network architecture inspired by the architecture proposed by Du Tran \emph{\textit{et al.}}\cite{tran2015learning}. The research by Du Tran \emph{\textit{et al.}}\cite{tran2015learning} originally used for action recognition in videos was extended utilizing a 2-D LSTM network to allow for face recognition in videos. We compared the performance of our architectures to existing state-of-the-art CNN architecture, Google/Inception\cite{schroff2015facenet}. The comparison is performed using the YouTube Faces database with both contrastive loss and triplet loss functions for similarity learning. The experiments show that the Google/Inception architecture with contrastive loss outperformed in top $n$ rank face retrievals when using a basic cosine distance on the learned feature representations. However, feature representations learned from the 2-D LSTM with triplet loss were superior to others for $n$-shot facial identification. These learned features were evaluated using a non-linear Support Vector Machine (SVM) and compared to further techniques. Therefore, we have demonstrated there are scenarios where early data fusion of spatial and temporal features are useful in increasing performance in face recognition tasks.

\section{Background and related works}
\subsubsection{Transfer Learning with joint Bayesian} proposed by Cao \emph{\textit{et al.}}\cite{cao2013practical} used a transfer learning algorithm based on the joint Bayesian method for face recognition. That is they proposed the transferring of a model trained on abundant data in one domain to create a classifier that performed ideally as well in the unseen data. Kullback-Leibler divergence was used as a regularizer for the objective function and was optimized using a generalized expectation-maximization algorithm. Cao \emph{\textit{et al.}}\cite{cao2013practical} reported a performance of $96.33\%$ on the labeled faces in the wild dataset.

\subsubsection{The Triplet sampling} method for similarity learning was used by Wang \emph{\textit{et al.}}\cite{wang2014learning}. The paper presented a deep ranking model with triplet sampling and uses deep learning to characterize fine-grained image similarity for classification. The network takes image triplets as input and each image is fed independently into three identical neural networks that share identical architecture and weights. The output of the neural network layer is fed into the ranking layer where it evaluates the hinge loss of the image triplet. For optimization, a distributed stochastic gradient algorithm with momentum was implemented and a triple sampling strategy was employed to select the optimal triplet from the data set. It was reported that the deep ranking model achieved $85.7\%$ for information retrieval on the test dataset.

\subsubsection{Facenet} is a model developed by Google that uses 128-dimensional representations of faces generated by an extremely deep neural network \cite{schroff2015facenet}. The model was trained with 260 million images using triplet loss. An online triplet sample selection strategy was used to generate triplets from the dataset to train the model. The model was reported to perform $99.63\%$ on the labeled faces in the wild dataset.

\subsubsection{Deep Learning Face Representation} proposed by Sun \emph{\textit{et al.}}\cite{sun2014deep} as a way of learning high-level features using Deep CNN, DeepID, for face verification by learning $10000$ unique identities. Facial keypoints were detected using a facial point detection method\cite{sun2013deep}. All faces from the images were aligned using a similarity transform for consistency. Features were then extracted from the patches and regions around the facial keypoints to train a deep CNN to extract the DeepID feature representations of each face. A joint Bayesian model was used for face verification. From experiments it is demonstrated that a key point feature representations combined with the Joint Bayesian model increased accuracy linearly as the unique faces increased. The authors claimed that with transfer learning Joint Bayesian coupled with DeepID achieved $97.45\%$ accuracy on the labeled face in the wild dataset.

\subsection{CNN architectures}
   \subsubsection{GoogLeNet/Inception}
GoogLeNet/Inception is a deep CNN architecture developed by Szegedy \emph{\textit{et al.}}\cite{szegedy2015going}. The primary contribution of the architecture was providing a novel way of improving the performance of the deep neural networks beyond relying on increased depth as proposed in \emph{\textit{Simonyan et al.}}\cite{simonyan2014very}. The drawback of the larger networks is they consist of a more sizeable number of parameters, which could lead to over-fitting and an increase in computational resources. Consequently, the Inception architecture was proposed to provide improved utilization of computing resources in the network by reducing the number of parameters. The architecture allows the depth and width of the network to increase while managing computational resources. This resource management is achieved through careful design of "Inception modules" that apply dimensionality reduction in the network wherever the computational requirements are excessively high.
\subsection{Convolutional 3D (C3D)}
C3D is a deep 3-dimensional CNN architecture developed by Du Tran \emph{\textit{et al.}}\cite{tran2015learning} for the purpose of learning spatio-temporal features in sequences of images (videos). The paper contributed towards action recognition in videos by finding that 3-dimensional convolutions are more capable of learning spatio-temporal features compared to 2-dimensional convolutions. The paper also showed that a $3\times3\times3$ convolution kernels for all layers is the preferred setup for the 3-Dimensional ConvNets. The proposed architecture outperformed the Imagenet baseline in action recognition on the CIFAR101 videos dataset.
    
    \begin{figure*}[ht]
      \includegraphics[height=18cm,angle=90]{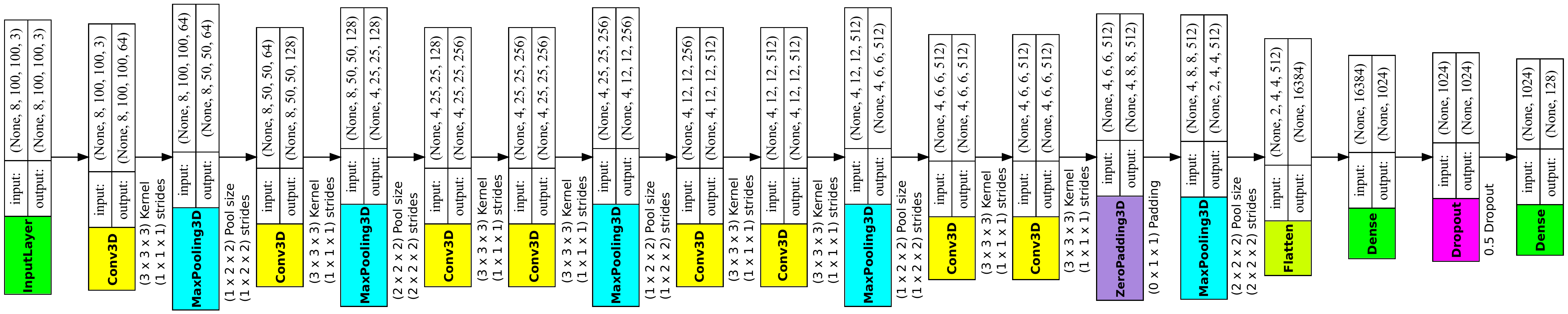}
      \caption{C3D architecture used in our implementation. The pool sizes of the third and fourth MaxPooling3D layer was modified. One less dense and dropout layer compared to previous C3D \cite{tran2015learning}. The value of the second last dense layer was modified from $4096$ to $1024$ and the last dense layer has output size of 128.}
      \label{fig:C3D}
    \end{figure*}
    
    \subsection{2-D LSTM}
Long short-term memory (LSTM) represents a type of recurrent neural network architecture that makes use of memory cells to store and feedback information at each time step of the inputs. LSTMs are capable of learning and better finding long-range temporal relationships in time series data like videos. A paper by Yue \emph{\textit{et al.}}\cite{yue2015beyond} proposed the use of learning short snippets of videos by passing the frames of the snippets into a CNN to produce sequences of embeddings. The sequences of embeddings are used as input into a 5-layer LSTM architecture to learn action recognition on CIFAR-101 videos dataset. We have utilized a 5-layer 2-D LSTM architecture for the purpose of face recognition in video.
    
    \begin{figure*}[ht]
      \includegraphics[height=18cm,angle=90]{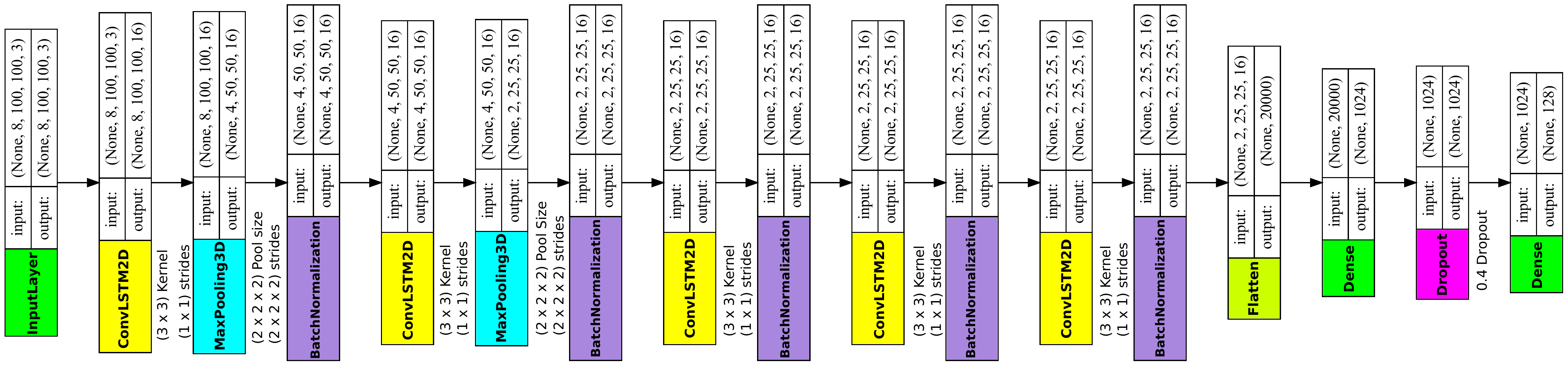}
      \caption{2-D LSTM architecture used in our implementation.}
      \label{fig:LSTM}
    \end{figure*}
    
\section{Methodology}
The purpose of this research is to make use of existing metric learning techniques to train similarity learning models. The similarity learning models are used for facial recognition, as well as similar image retrieval. The research aims to assess the models and confirm whether they provide satisfactory performances on unseen real-world data. We have trained CNNs with different similarity learning techniques as well as baseline models to provide performance comparisons.

\subsection{Dataset} 

The dataset used for our experiment was the YouTube Faces Database based on Labeled Faces in the Wild (LFW) identities originating from the paper by Wolf \emph{\textit{et al.}}\cite{wolf2011face}. The database was acquired by taking each identity from LFW and searched YouTube to get the top six video clips for each identity. The outcome from the YouTube search yielded a total of 3425 videos for 1595 unique subjects. From the video samples we extracted a total of 621126 frames, where each frame contains a single face. All the frames of the dataset were resized to $100 \times 100 \times 3$ dimensions. The total size of the dataset was 2.4 Gigabytes (GB). 


\subsection{Training set}
The training set for Google/Inception network architecture consists of videos from 1276 randomly chosen identities ($80\%$) from the total 1595 identities and the videos were split into frames. The faces were cropped from the frames and any unnecessary background information was removed. The cropping coordinates were obtained using a Viola-Jones detector. The training set for C3D and 2-D LSTM architecture consists of $1276$ randomly chosen identities with the same seed as the training set of Google/Inception Network. The videos are segmented into sequences of eight images and any remaining sequences with fewer frames are ignored. 

\subsection{Validation Set}
The still image validation set for Google/Inception Network was obtained by splitting $20\%$ of the training set at a random seed. The validation split for sequences of images was $33\%$ of the training data. The larger percentage was to ensure there are at least two sequences for each identity to allow us to generate pairs and triplets we required.

\subsection{Testing Data}
The test data used consisted of the remaining 319 identities ($20\%$) of the dataset with dimensions identical to the training and validation sets. Still images were employed to test the Google/Inception Network and sequences of eight image clips were used for testing C3D and the 2-D LSTM networks. The out-of-sample identities enabled us to evaluate how well the models will generalize to unseen identities.

\subsection{Pairs and Triplets}
The pairs and triplets for training and validation were generated by randomly sampling $16$ unique identities. Two images/sequences were taken each of the $16$ identities to ensure we are able to make pairs and triplets. Random positive and negative pairs were generated and random triplets were mined with semi-hard negative online mining. A variety of diverse identities during training allows us to optimize distances between positive and negative classes. 

\subsection{Architecture of CNNs}
The architecture implemented for our image CNN was the GoogleNet architecture also known as InceptionV1. The architecture has won the ImageNet Large Scale Visual Recognition Challenge 2014 competition which means that it is a credible architecture to select for our task\cite{szegedy2015going}. The architecture implemented for our sequence image data will be the C3D and 5-layer stacked 2-D LSTM described above. They have both previously been employed for action recognition and sports classification. The C3D and 2-D LSTM had yet to use with contrastive and triplet loss for face recognition in videos. We compared the performance of short sequence learning for face recognition and still images to investigate if there are any benefits to using short image sequences for facial recognition.  

\subsection{CNN setups}
The Google/Inception architecture used in our experiment was set up the way as described in the paper by Szegedy \emph{\textit{et al.}}\cite{szegedy2015going} with the input dimension of $100 \times 100 \times 3$ and the output feature representation of the size $128$ for still images. \newline
The C3D architecture implemented was setup as described in the paper by Du Tran \emph{\textit{et al.}}\cite{tran2015learning}. The difference being the final classification layer was replaced with a representation embedding layer. The representation layer contained $128$  features and was linearly activated similar to the Google/Inception architecture\cite{szegedy2015going}. The input layer of the network was dimensions of $8 \times 100 \times 100 \times 3$ for video sequences. The layout of the architecture can be seen in Fig \ref{fig:C3D}.
\newline
The 2-D LSTM was set up with an input layer of $8 \times 100 \times 100 \times 3$ followed by a 2-D LSTM layer with 16 filters. The input layer was followed by a pooling layer and finally batch normalization layer. The LSTM layer, pooling layer, and batch normalization was repeated five times followed by a flatten layer and a dense layer. The output layer was a linearly activated representation embedding layer of $128$ like the Google/Inception architecture\cite{szegedy2015going}. The network contains five 2-D LSTM layers, and the layout is presented in Fig \ref{fig:LSTM}.
\subsection{Triplet loss}
For the model to learn optimal feature representations, we need a suitable loss function. We selected the triplet loss function proposed in the paper by Wang \emph{\textit{et al.}}\cite{wang2014learning} as it was able to learn good feature representations to determine how similar images are. The triplet loss function makes use of triplets of images. The triplet consists of an anchor ground truth image, a positive image belonging to the same identity as ground truth, and a negative image from an identity that differs from the ground truth. The algorithm attempts to bring the positive image closer to the anchor and push the negative image away. The degree to which the negative is pushed is determined by a Euclidean distance margin to the anchor image. The loss function is defined as
\begin{equation}
    \mathcal{L} = \max(d(a,p) - d(a,n) + \text{margin},0),
\end{equation}

where $d$ represents euclidean distance, $a$ is the anchor ground truth image, $p$ is the positive image, $n$ is the negative image. The margin represents a radius around the anchor image. The function tries to optimize the difference between the anchor-positive and anchor-negative at the same time.

\subsection{Contrastive Loss}
The contrastive loss function finds optimal features by using pairs of images consisting of positive matching pairs and negative non-matching pairs. The loss function differs from triplet loss in that it tries to minimize the distance between positive pairs and maximize negative pairs in separate steps. The contrastive loss function is defined as
\begin{equation}
   \mathcal{L} = \frac{1}{2}(1-Y)(d(\tilde{y}_i, \tilde{y}_j))^2+ \frac{1}{2}(Y)\{\max(0,\text{margin}-d(\tilde{y}_i, \tilde{y}_j))\}^2,
\end{equation}
where $Y$ represents the label 0 or 1 and is 0 if the input is from the same class and 1 otherwise. $d(\tilde{y}_i, \tilde{y}_j)$ represents the Euclidean distance between the output feature representations of the network for the pair of images.

\subsection{Experiment Setup}

\subsubsection{Preprocessing}
Every image for Google/Inception network was cropped around the faces and resized to dimensions of $100 \times 100 \times 3$. The images were normalized by dividing by $255$ and the mean image was calculated and subtracted from the training, validation, and test set. The short sequences for C3D and 2-D LSTM was cropped around the faces and resized to dimensions of $8 \times 100 \times 100 \times 100 \times 3$. The sequences were normalized by dividing by $255$ for each frame of the sequences.

\subsubsection{Hyper-parameter searching}
Training a network using contrastive and triplet loss functions is a non-trivial task. It is extremely sensitive to the margin employed to bring images/sequences of the same class close and push those of different classes apart. The margin decides how far a negative image should be from a positive image and would lead to sub-optimal separation of identities with a randomly guessed margin. Therefore, it is critical to find the optimal margin for each of the deep learning setups. We at random selected $200$ identities from the original YouTube Faces Database and split the images/sequences into a train and validation set. The margins consisted of $0.25,0.5,0.75,1.0,1.25$ at the initial search for the optimal margin. We gradually narrowed it down depending on the top 1 rank retrieval accuracy on the validation set. If the margin was outside the initial margins, we would increase it by $0.25$. We half the margin if it is on the inside of the initial margins. The Adam optimizer was used during hyper-parameter searching with a learning rate of $0.0001$.

\subsubsection{Training Setup}
The Google/Inception, C3D, and 2-D LSTM networks were constructed with Python using the Keras Library with contrastive and triplet loss functions. To train the Siamese Network and Triplet Loss networks, we used batch sizes of 32 with 16 identities per batch and two images/sequences per identity. There are $16$ positive pairs and $16$ random negative pairs for Siamese Network and Triplet Loss Network employed semi-hard negative mining on each batch with 32 images/sequences. The Adam optimizer was used from Keras with a learning rate of $0.0001$, beta 1 value of $0.99$, and beta 2 value of $0.99$ across the various network architectures. 
\newline \newline
The hyper-parameter search yielded the optimal margins to be used for the Siamese and Triplet Loss networks. The margins that were used for our Siamese network setups were $[1.0,0.125,0.125]$ for the Google/Inception, C3D, and LSTM respectively. For our Triplet Loss networks, the margins employed were $[1.25,1.5,2.0]$ for the Google/Inception, C3D, and LSTM respectively. All the networks were trained until its loss converged measured with the validation set. The default epochs used were $25$ but was extended when necessary during our training. Each of our networks was trained five times to obtain an average performance using five random seeded splits of the identities for training and testing. The output of the networks was feature representations of size $128$ for each image/sequence into our network that we will use to measure top $n$ retrieval performance.

\subsection{n-Shot learning}
The concept of n-shot learning is making use of a scarce amount of data to train a machine to perform tasks such as object classification. The n represents the number of examples we have available of each class/category to learn from. There has been previous work that demonstrated the use of few or zero-shot learning capable of recognizing, classifying, and retrieving new and old objects classes \cite{reed2016learning,sung2018learning}. The previous models were able to learn good feature representations for performing those tasks implementing their unique architectures. In our research, we made use of limited feature representations for each of the identities. The feature representations are from the Google/Inception, C3D, and 2-D LSTM. We utilize the features to train an SVM classifier to perform identification on unseen data.

\subsection{Support Vector Machine for n-shot learning}
A support vector machine (SVM) was used to measure how useful the feature representations of each of the networks were for face identification. Hoffer \emph{et al.} (2015) have previously demonstrated the use of an SVM with feature representations learned from a deep convolutional network with triplet loss to perform classification on the MNIST dataset\cite{hoffer2015deep}. The SVMs were given only $n$ random samples of each individual from each of the networks to complete face identification on the rest of the samples. The experiment was performed on the test set with $200$ unique identities. The SVM was setup up with a linear kernel, max iteration of $2000$, and a tolerance of $1e^{-6}$. We found the optimal regularization parameters for each of the SVMs by searching through various intervals using GridSearchCV from Scikit-learn\cite{scikit-learn} library. The optimal regularization parameters for the SVMs trained with Google/Inception, C3D and 2-D LSTM with contrastive loss feature representations were $[0.001,0.001,10]$ respectively. The other optimal regularization parameters for the SVMs trained with Google/Inception, C3D and 2-D LSTM with triplet loss feature representations were $[10,10,0.1]$ respectively. A linear kernel was used since there exists literature that supports its use over a Radial Basis Function (RBF) when the data have high dimensionality \cite{hsu2003practical}. The feature representations of our data are of size $128$ which is moderately high.

\subsection{Hardware and software}
The hardware configurations that we have used for our research experiments consist of two machines. Machine one contains an AMD RYZEN 3600 processor, 16 GB of RAM, and a GTX 1060 6GB GPU and machine two contains an AMD RYZEN 3700X processor, 16 GB of RAM and an RTX 2070 8GB GPU. Both machines used Linux operating systems, the python version was 3.6, the Keras version was 2.2.4\cite{chollet2015keras} with Tensorflow-GPU backend version 1.14.0 \cite{tensorflow2015-whitepaper} and Scikit-learn\cite{scikit-learn} software kit was used in our implementations.

\section{Results and Discussion}
In this section, we will be discussing the testing metric and the testing procedure that was used to acquire results for each of the models. The models consist of the Siamese and the Triplet Loss networks with Google/Inception, C3D, and 2-D LSTM architectures.

\subsection{Top n Rank Retrievals with cosine similarity}
We used top $n$ retrievals to measure the performance of each of the networks on the test set. We obtain the feature representations of the test set on each of the networks. Each sample in the test set was used as a query image. Its feature representation was compared to all other samples in the test set using cosine similarity. This comparison was performed to measure the distance between each of the other samples. Top $n$ retrieval accuracy is measured by taking $n$ closest samples from the query results. The retrieved samples are evaluated as being relevant to the query which in our case means that the result's labels are the same as the query's label.

\subsection{Siamese and Triplet Loss Top $n$ Retrieval Results}
\begin{table}[H]

\label{tab:ContrastiveTable}
\caption{Retrieval Accuracy of the Three different architectures with Contrastive loss function}
\begin{tabular}{|c|c|c|c|} \hline
     & Top 1 & Top 3 & Top 5 \\ \hline
Google/Inception            &    $\textbf{99.6}\pm \textbf{0.06\%}$ & $\textbf{99.71}\pm \textbf{0.04\%}$  &  $\textbf{99.75}\pm \textbf{0.03\%}$      \\ \hline
C3D                         &    $98.89\pm 0.19\%$ & $99.26\pm 0.08\%$   & $99.38\pm 0.07\%$        \\ \hline
2-D LSTM    &    $91.42\pm 1.57\%$ & $93.79\pm 1.17\%$ &  $94.7\pm 1.02\%$       \\ \hline
\end{tabular}
\end{table}
The results above show Siamese network with Google/Inception architecture with contrastive loss outperforms the C3D and 2-D LSTM in terms of top $n$ rank retrievals. All three models provide excellent performance. It is also shown that there are no benefits to using sequences as compared to still images of the videos for Siamese Networks with contrastive loss when comparing the three Architectures. The results were averaged over five numerous train, validation, and test splits. It was demonstrated that C3D and 2-D LSTM with contrastive loss function was not optimal for metric learning as compared to a still image architecture like Google/Inception.
\begin{table}[H]

\label{tab:TripletTable}
\caption{Retrieval Accuracy of the Three different architectures with Triplet loss function}
\begin{tabular}{|c|c|c|c|} \hline
                            & Top 1 & Top 3 & Top 5 \\ \hline
Google/Inception            &   $85.04\pm 1.93\% $ & $90.53\pm 1.27\%$  & $92.47\pm 1.01\%$       \\ \hline
C3D                         &    $85.43\pm 1.78\%  $ &  $91.29\pm 1.14\%$ &  $93.2\pm 0.87\%$      \\ \hline
2-D LSTM    &  $\textbf{98.51}\pm \textbf{0.12\%}$&  $\textbf{99}\pm \textbf{0.07\%}$    &  $\textbf{99.15}\pm \textbf{0.07\%}$        \\ \hline
\end{tabular}

\end{table}
The Triplet Loss network results provide a contrast to the results from the Siamese network for top $n$ rank retrievals. The results show that Google/Inception with triplet loss performs the worst on average compared to the C3D and 2-D LSTM with triplet loss. The 2-D LSTM performs significantly better overall. Siamese Networks' results are better overall compared to the triplet loss results on the YouTube Faces Dataset. However, there are benefits to using C3D and 2-D LSTM with triplet loss compared to Google/Inception with triplet loss. Our results contradict an outcome from the person re-identification literature\cite{hermans2017defense}. In this previous work it was demonstrated that a ResNet-50 architecture with triplet loss outperformed a Gate Siamese Network with contrastive loss in top $n$ retrieval performance. This result defends the re-evaluation of triplet loss for this task. The comparison of these two loss functions in the literature\cite{hermans2017defense} was not consistent due to the two different Architectures. The contrast in results between Table I and Table II indicates that C3D and 2-D LSTM were more optimally combined with the triplet loss function. Decidedly, they do not benefit from the use of contrastive loss function.

\subsection{SVM with n-shot learning results}
\begin{table}[H]
\label{tab:SVMTable}
\caption{Accuracy of the Three different architectures with SVM for face identification}
\begin{tabular}{|c|c|c|c|c|c|c|} \hline
n & \MyHead{0.9cm}{Google/ \\ Inception contrastive loss} & \MyHead{0.9cm}{C3D contrastive loss} & \MyHead{0.9cm}{2-D LSTM contrastive loss} & \MyHead{0.9cm}{Google/ \\ Inception triplet loss} & \MyHead{0.9cm}{C3D triplet loss} & \MyHead{0.9cm}{2-D LSTM triplet loss}\\ \hline
1    & $ 26.83\%$                  & $5.44\%$           & $37.24\%$     & $27.85\%$   & $42.37\%$ & $\textbf{51.61\%}$\\
2    & $ 27.04\%$                  & $1.14\%$          & $36.90\%$      & $30.46\%$    & $52.79\% $  & $\textbf{65.91\%}$\\
3    & $ 29.44\%$                   & $1.75\%$           & $38.24\%$    &  $33.43\%$   & $57.29\%$ & $\textbf{73.44\%}$\\
4    & $ 31.22\%$               & $1.78\%$             & $38.59\%$      &  $36.84\%$  & $60.84\%$ & $\textbf{78.79\%}$ \\
5    & $ 30.03\%$               & $2.02\%$                & $37.88\%$    &  $38.49\% $     &  $63.56\%$ & $\textbf{82.30\%}$\\
6    & $31.10\%$              & $4.04\%$               & $38.05\%$   &  $39.27\%$   & $66.25\%$ & $\textbf{84.44\%}$\\
7    & $30.86\%$               & $4.23\%$                & $38.71\%$   &  $41.12\%$   & $68.57\%$ & $\textbf{86.80\%}$\\
8    & $30.89\%$              & $4.70\%$                 & $39.32\%$    & $41.41\%$   & $70.68\%$ & $\textbf{88.38\%}$ \\
9    & $30.66\%$                   & $4.50\%$              & $38.92\%$      &  $42.12\%$  & $72.09\%$ & $\textbf{88.95\%}$\\
10   & $30.49\%$                  & $6.64\%$             & $39.84\%$    &  $43.37\%$   & $73.66\%$ & $\textbf{90.20\%}$\\ \hline
\end{tabular}
\end{table}
The results in Table III show that the feature representations from the 2-D LSTM with triplet loss function combined with an SVM delivered the most impressive general performance. This performance was measured in terms of facial identification with $n$ samples given of each identity. The C3D architecture with triplet loss was second overall and shows the C3D and 2-D LSTM architecture provided good representation learning capabilities. C3D with contrastive loss gave the worst face identification performance overall with an SVM. 
\\
The Google/Inception with contrastive loss performed the best overall with top $n$ rank retrievals. This result can be observed in Table I when query images were compared to the entire test set of images to retrieve the closest image for facial recognition. With limited known samples from each identity to perform face recognition on the test set, we show that the feature representations from the CNNs with contrastive loss are not as good as the feature representations from the CNNs with triplet loss. This result is shown by the SVM results in Table III. The results obtained are in line with the results from the literature\cite{hoffer2015deep}. Here, it was demonstrated that a deep learning model with triplet loss provided better feature representations for face identification with SVMs compared to a deep learning model with contrastive loss function on the MNIST dataset.

\section{Conclusions}
In this research, we investigated the use of 3-D CNN and 2-D LSTM network architectures that are capable of learning temporal features from video sequences together with similarity learning. Similarity learning methods included contrastive and triplet loss for face recognition. We compared the performance of those networks to an existing state-of-the-art Google/Inception architecture with contrastive and triplet loss that learns from still images of videos. C3D and 5-layer stacked 2-D LSTM have been used for action recognition in videos in the past \cite{tran2015learning,yue2015beyond}.  In this research, we modified these architectures to support facial recognition in videos. The results show that the Google/Inception architecture with contrastive loss outperforms all the other network setups. Google/Inception's outperformance was in the top $n$ retrieval task indicating there are no immediate benefits to considering temporal features from video sequences for that use case. However, C3D and 2-D LSTM with triplet loss both outperformed Google/Inception when combined with triplet loss. Using SVM, we show that CNNs with triplet loss learned better feature representations compared to contrastive loss for face identification when we have only limited samples for each identity. The top two performances were from the 2-D LSTM and C3D with triplet loss which indicates there are benefits from learning temporal features from video sequences for face recognition. Further, it indicates we should explore other ways of fusing the temporal aspects of sequences when doing recognition tasks beyond the 2-D features.

\bibliography{annot}
\bibliographystyle{plain}

\end{document}